*Article*

# Method to Characterize Potential UAS Encounters Using Open Source Data

Andrew Weinert *

MIT Lincoln Laboratory, Lexington, MA, 02420-9176, USA ; andrew.weinert@ll.mit.edu

**Abstract:** As unmanned aerial systems (UASs) increasingly integrate into the US national airspace system, there is an increasing need to characterize how commercial and recreational UASs may encounter each other. To inform the development and evaluation of safety critical technologies, we demonstrate a methodology to analytically calculate all potential relative geometries between different UAS operations performing inspection missions. This method is based on a previously demonstrated technique that leverages open source geospatial information to generate representative unmanned aircraft trajectories. Using open source data and parallel processing techniques, we performed trillions of calculations to estimate the relative horizontal distance between geospatial points across sixteen locations.

## 1. Introduction

The continuing integration of unmanned aerial system (UAS) operations into the National Airspace System (NAS) requires new or updated regulations, policies, and technologies to maintain safety and enable efficient use of the airspace. One enabling technology to help address several UAS airspace integration gaps are airspace encounter models, which have been fundamental to quantifying airborne collision risk for manned and unmanned operation [1–5]. These models represent how aircraft behavior and their relative geometries evolve during close encounters. They have supported the development of surveillance and communication requirements [6,7].

*1.1. Motivation*

Mitigations for airborne collision risk and optimization of airspace operations are strongly dependent on the distribution of geometries and behavior of aircraft encounters. For example, collision avoidance systems are designed to determine, communicate, and coordinate avoidance maneuvers when they determine that a maneuver is needed to avoid a collision. These systems are the last and third layer for airspace conflict management and are employed after separation provision and strategic mitigation have failed. Collision avoidance systems may leverage vehicle-to-everything (V2X) communication technologies to improve performance and safety. How an aircraft behaves will influence the development of V2X routing protocols, link budgets, and energy requirements [8].

Fast-time Monte Carlo simulations are often utilized to evaluate the performance of aviation safety systems, such as detect and avoid (DAA) for UAS, for close encounters between aircrafts [2–4]. The design and effectiveness of these simulations are dependent on how close encounters are defined. The relative geometries and separation between aircraft are important criteria when defining these encounters. Simulated encounters are generated based on what safety function is being evaluated and the performance or behavior of the aircraft involved. For example, encounters meant to prompt collision avoidance maneuvers may not be suitable to evaluate a system attempting to maintain well clear.

Encounters and manned aircraft behavior have historically been based on an abundance of radar observations, with these models refined for decades [1–3]. However, as UAS are not routinely operating beyond visual line of sight, we cannot characterize UAS encounters using observed aircraft behavior. While independent trajectories of representative UAS behavior can be generated [2,3], an alternative approach to characterize the geometry and feasibility of potential UAS vs. UAS encounters is required for the development and evaluation of DAA safety systems.

*1.2. Scope*

We adopted a similar scope as the ASTM standard [9] and the Federal Aviation Administration (FAA) proposed [1] UAS Remote Identification (Remote ID) rule for considering close airborne encounters involving UASs. Out of scope was estimating the frequency of potential encounters and non-DAA concerns, such as airworthiness. This scope was informed by the FAA UAS Integration Office, and the standards activities of International Civil Aviation Organization (ICAO), ASTM F38, RTCA Special Committee 147, and RTCA Special Committee 228. The Airborne Collision Avoidance System (ACAS X) unmanned variants [10,11] were considered as DAA reference architectures, as they are being standardized by the aforementioned organizations.

The scope encompassed commercial and recreational UASs weighing greater than 0.55 pounds with no restrictions on UAS size, performance, or operating altitude. We assumed that encounters consist of only two aircrafts, but that the general policy for manned formation flight will apply to UASs. We prioritized research to DAA and collision avoidance systems designed to mitigated the risk of midair collision. These prioritized system are the third and final layer for airspace conflict management and are employed after separation provisions and strategic mitigates have failed [12].

*1.3. Objectives and Contribution*

We focused on one of the many objectives identified by the aviation community to support integration of UASs into the airspace, quantitatively define a close encounter between UASs. Satisfying this objective could lead to or enhance existing risk-based minimum operating performance standards (MOPS) for a DAA system [12]. In response, the primary contribution was an analytical method that uses freely available open source data to estimate range and azimuth between potential UAS operations. We applied this method to calculate the distance between various realistic short and medium term UAS inspection operations. We focused on inspection of long linear infrastructure and point obstacles due to the industry demand and to demonstrate that a variety of data sources can be used with this method. The results can easily be extended with other use cases or enhanced using digital elevation models to estimate potential vertical separation.

These results directly informed the scoping and generation of simulated Monte Carlo encounters of two UASs to support DAA development within RTCA SC-147[2]. However, discussion of these Monte Carlo encounters were out of scope for this paper. Additionally, while DAA systems can be obstacle aware [10], our research does not directly support a UAS capability to avoid obstacles. Rather, these results can inform if obstacles should be represented as part of simulated UAS encounters when designing or evaluating DAA capabilities to mitigate the risk of a midair collision between aircraft.

We intend to release the software used for this analysis under a permissive open-source license. These contributions are intended to support current and expected UAS DAA system development and evaluation, specifically estimating the probabilities associated with encountering a low-altitude aircraft based on geography or defining total allowable systems latency of a safety critical system that satisfies performance-based requirements. The results inform how to generate representative UAS trajectories [3] and pair trajectories together to simulate encounters [2]. This paper is complimented by another effort to propose a quantitative metric to assess the performance of a smaller UAS safety system [8,9].

---

[1] https://www.regulations.gov/docket?D=FAA-2019-1100

[2] https://www.rtca.org/sc-147/

## 2. Materials and Methods

Our experiment was based on calculating the distance between any given points along different surveillance or inspection targets for a UAS. The experiment design was based on an approach to generate representative UAS trajectories that take into account their operational intent by leveraging open source datasets, such as OpenStreetMap (OSM), "a knowledge collective that provides user-generated street maps [13]." For pairs of latitude and longitude coordinates from different features, we calculated the relative geometry between them. This contrasts an alternative approach of generating representative UAS trajectories based on open source features and characterizing encounters between these trajectories in a six degree of freedom simulation. The closed form analytical approach is similar to what Edwards and MacKay used to determine surveillance requirements for UAS DAA [7].

All data used were freely and easily accessible from the internet and the software we developed for this analysis has been released under a permissive open-source license. Our described method and subsequent results are reproducible, given access to the appropriate MATLAB toolboxes.

### 2.1. Use Cases and Data Sources

We evaluated use case pairs for sixteen locations. These areas include all locations associated with the UAS Integration Pilot Program (IPP)[3], majority of states with FAA UAS test sites[4], a few states within FEMA Region 1, and the territorial island of Puerto Rico. All geospatial data used were freely sourced from the public domain. For locations, their administrative boundaries were sourced from Natural Earth Data at a 1:10m scale. The analysis focused primarily on UAS inspections as long linear infrastructure[14–16] and point obstacles. These realistic use cases are some the most mature and well understood UAS operations, as indicated by their evaluation as part of the UAS IPP, FAA UAS test site, or FAA Pathfinder activities.

Point features were sourced from the FAA digital obstacle file (DOF)[5] and locations of 6300 wind turbines were sourced from the United States Wind Turbine Database[17]. While the FAA DOF may include wind turbines, we wanted to assess the sensitivity of the results between a general (FAA DOF) and a specific dataset (USWTDB). Specifically, the FAA DOF had 483,279 obstacles but we only considered the 321,699 obstacles that were at least 50 feet tall. The software to parse the FAA DOF has been released under a permissive open source license[6].

Electric power transmission lines operating at high voltages of 69-765 kV were sourced from the U.S. Department of Homeland Security (DHS) Homeland Infrastructure Foundation Level Data (HIFLD)[7]. Regular railway tracks were sourced from the Geofabrik OSM extracts[8]. Four types of pipelines (crude oil, hydrocarbon gas liquids, natural gas, and petroleum products) were sourced from the U.S. Energy Information Administration (EIA)[9]. Federal and state freeways and primary roads were sourced from Natural Earth Data[10]. While electric power lines, railways, and roads are mostly above ground, the majority of oil pipelines are buried underground. This results in relatively straighter infrastructure layouts.

Polygons of golf course perimeters were also sourced from the Geofabrik OSM extracts as a representative recreational feature. Other recreational features such as beaches and lakes were considered but ultimately not included to manage computation resources. Agricultural land use features such as farms and vineyards were not included because we heuristically assessed that OSM did not have sufficient coverage and we could not find an appropriate alternative dataset with nationwide coverage.

---

[3] https://www.faa.gov/uas/programs_partnerships/integration_pilot_program/

[4] https://www.faa.gov/uas/programs_partnerships/test_sites/

[5] https://www.faa.gov/air_traffic/flight_info/aeronav/digital_products/dof/

[6] https://github.com/Airspace-Encounter-Models/em-core/tree/master/matlab/utilities-1stparty/faadof

[7] https://hifld-geoplatform.opendata.arcgis.com/datasets/electric-power-transmission-lines

[8] https://download.geofabrik.de/north-america.html

[9] https://www.eia.gov/maps/map_data/CrudeOil_Pipelines_US_EIA.zip

[10] https://www.naturalearthdata.com/downloads/10m-cultural-vectors/roads/

Line vector and polygon features were interpolated to enforce a consistent spacing between points. Interpolation was calculated based on the arc length between points along the vector using a linear chordal approximation. This was more efficient than using a piecewise cubic Hermite interpolating polynomial (pchip) used by previous modelling efforts [1]. The spacing of 500 feet was selected because it preserves details along curves and was equivalent to the 500 feet horizontal dimension of the near mid-air collision (NMAC) safety metric, which is commonly used to evaluate airborne collision risk. This spacing also heuristically determined to be sufficiently efficient, as a linear reduction in spacing results in a non-linear increase in computation time. This trade-off is further discussed in Section 4.1.

*2.3. Processing and Workflow*

The workflow was organized into pre-processing, geometry calculations, and results aggregation; with each component encoded by a dedicated MATLAB script. First, pre-processing consisted of the following:

1. Download open source data for use cases
2. Filter data based on administrative boundaries
3. For FAA DOF, create small circle vectors centered on reports points with a radius of the horizontal position uncertainty
4. Interpolate data to have a fixed spacing of 500 feet between points
5. Aggregate all vectors into a single array of latitude and longitude points
6. Recheck and confirm that all points are within administrative boundary

Next, distance calculations were completed, organized by administrative boundary. A maximum distance of 60 nautical miles was specified to reduce the quantity of discrete distance computations. As the scope focused on close encounters, points greater than 60 nautical miles apart should not be considered "close" for aviation operations. The geometry calculates consisted of the following:

7. Calculate unique pairs of features (e.g., railways and roads)
8. Create a small circle with a 60 nautical mile radius centered on each point for one of the features
9. Identify which points of the other feature are within each small circle
10. Calculate the distance using the WGS84 reference ellipsoid between the center of the small circle and all point-in-polygons
11. Determine the closest point by calculating the minimum for all computed distances

After the geometry calculations, the results were aggregated across locations and pairs of features. General statistics such as the mean, median, and other percentiles were calculated.

*2.4. Azimuth*

Azimuth is the angular distance along a fixed reference direction to an object. The azimuth, also known as bearing, between two aircrafts is a geometric variable used when defining encounters between aircrafts. This angular measurement is often used when considering aircraft right-of-way rules regarding whether an aircraft should maneuver left or right to avoid a collision. In developing this analysis, we originally calculated azimuth in addition to distance between points. However, the aggregate azimuth distributions provided no justification to favor one specific relative orientation over another. Thus, we did not calculate angular distance as part of this analysis.

## 3. Results

There were 16 unique locations and 21 unique pairs of features. Across all 336 combinations of locations and pairs of features combinations, about $2.71 \times 10^{12}$ pairs of points were considered. High performing computing resources from the MIT Lincoln Laboratory Supercomputing Center [18] were leveraged for the trillions of discrete calculations. This section is organized into general statistics focused on average distance between points, followed by more detailed results using percentiles.

### 3.1. General Statistics

Tables 1 and 2 report the total pairs of points considered and the mean of the closest point of approaches for various subsets organized by location (Table 1) or use cases (Table 2). Table 3 aggregates pairs of features across all locations. The weights used for calculating the weighted mean were per row rather than across the entire table. Please refer to the source code for implementation details. Tables 1 and 2 support the following colloquial statements:

- "On average, any two features of interest are closer to each other in Massachusetts than Kansas;"
- "While Nevada is the 7th largest state by area and New York is the 27th, there are more potential ways for UASs to encounter each other in New York, given the features of interest;" and
- "On average, railways and roads are closer to each other than railways and wind turbines."

**Table 1.** General statistics for each location. Means are in nautical miles with extremes highlighted.

| ISO 3166-2 Code | Location | Total Point Pairs | Weighted Mean | Unweighted Mean |
|---|---|---|---|---|
| US-CA | California | $3.468 \times 10^{11}$ | 4.98 | 10.83 |
| US-FL | Florida | $8.387 \times 10^{10}$ | 3.60 | 18.58 |
| US-KS | Kansas | $9.718 \times 10^{10}$ | 4.06 | 8.63 |
| US-MA | Massachusetts | $6.176 \times 10^{9}$ | 2.06 | 3.94 |
| US-MS | Mississippi | $3.359 \times 10^{10}$ | 3.48 | 4.49 |
| US-NC | North Carolina | $4.591 \times 10^{10}$ | 4.15 | 20.00 |
| US-ND | North Dakota | $4.906 \times 10^{10}$ | 6.28 | 9.86 |
| US-NH | New Hampshire | $1.078 \times 10^{9}$ | 3.88 | 9.92 |
| US-NV | Nevada | $1.060 \times 10^{10}$ | 10.67 | 17.81 |
| US-NY | New York | $8.997 \times 10^{10}$ | 3.35 | 7.85 |
| US-OK | Oklahoma | $1.514 \times 10^{11}$ | 3.98 | 10.39 |
| US-PR | Puerto Rico | $1.611 \times 10^{8}$ | 2.71 | 8.58 |
| US-RI | Rhode Island | $9.362 \times 10^{7}$ | 1.68 | 3.67 |
| US-TN | Tennessee | $3.335 \times 10^{10}$ | 3.30 | 17.76 |
| US-TX | Texas | $1.722 \times 10^{12}$ | 4.54 | 11.71 |
| US-VA | Virginia | $3.476 \times 10^{10}$ | 3.07 | 3.29 |

**Table 2.** General statistics for feature pairs. Means are in nautical miles with extremes highlighted.

| Feature #1 | Feature #2 | Total Point Pairs | Weighted Mean | Unweighted Mean |
|---|---|---|---|---|
| FAA Obstacles | Golf Course | $1.519 \times 10^{10}$ | 7.89 | 6.87 |
| FAA Obstacles | Pipeline | $2.290 \times 10^{11}$ | 2.22 | 5.76 |
| FAA Obstacles | Railway | $8.933 \times 10^{10}$ | 6.01 | 4.73 |
| FAA Obstacles | Road | $5.762 \times 10^{10}$ | 4.00 | 2.89 |
| FAA Obstacles | Wind Turbine | $4.372 \times 10^{9}$ | 16.36 | 23.28 |
| Electric Power | FAA Obstacles | $2.281 \times 10^{11}$ | 2.00 | 1.96 |
| Electric Power | Golf Course | $6.537 \times 10^{10}$ | 7.84 | 7.94 |
| Electric Power | Pipeline | $7.532 \times 10^{11}$ | 2.47 | 6.43 |
| Electric Power | Railway | $3.492 \times 10^{11}$ | 5.87 | 5.43 |
| Electric Power | Road | $2.241 \times 10^{11}$ | 3.92 | 3.61 |
| Electric Power | Wind Turbine | $1.525 \times 10^{10}$ | 28.57 | 28.73 |
| Golf Course | Pipeline | $3.680 \times 10^{10}$ | 2.45 | 5.22 |
| Golf Course | Railway | $2.540 \times 10^{10}$ | 3.28 | 3.88 |
| Golf Course | Road | $1.674 \times 10^{10}$ | 2.22 | 2.35 |
| Golf Course | Wind Turbine | $8.258 \times 10^{8}$ | 28.41 | 29.05 |
| Pipeline | Railway | $2.890 \times 10^{11}$ | 7.72 | 5.08 |
| Pipeline | Road | $1.913 \times 10^{11}$ | 5.36 | 4.20 |
| Pipeline | Wind Turbine | $1.600 \times 10^{10}$ | 29.49 | 30.56 |
| Railway | Road | $8.914 \times 10^{10}$ | 2.17 | 2.35 |
| Railway | Wind Turbine | $5.898 \times 10^{9}$ | 28.64 | 28.26 |
| Road | Wind Turbine | $3.712 \times 10^{9}$ | 30.92 | 29.64 |

Figure 1 illustrates the weighted means given the number of point pairs for Tables 1 and 2. The weighted mean for most locations was less than five nautical miles. There was more variability when comparing across pairs of features. The minimum weighted average in Figure 1 was 1.68 nautical miles and the maximum was 30.92 nautical miles.

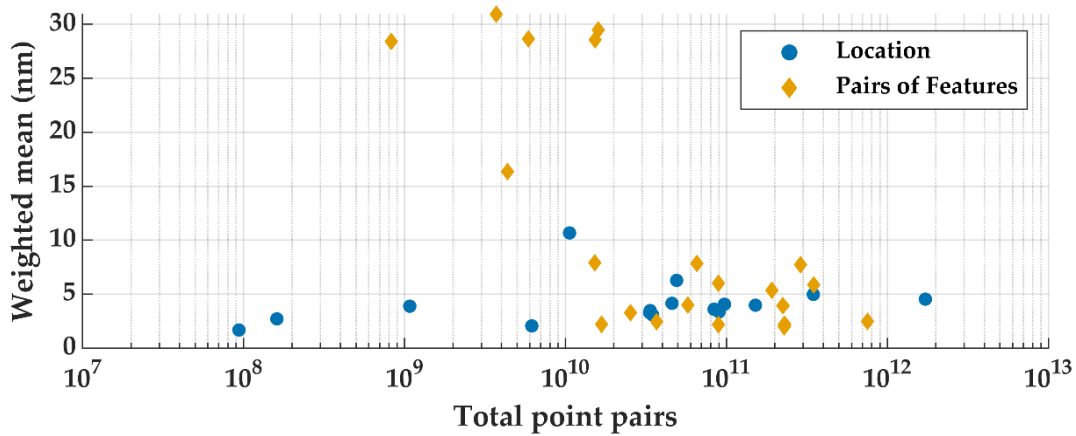

**Figure 1.** Illustrated weighted means for closest point of approach from tables 1 and 2.

*3.2. Percentiles*

A set of distance percentiles was calculated for each pairs of features for all locations. As an example, Table 3 provides the percentiles for California. Please refer to the supplemental material for the percentiles of all 336 combinations of locations and pairs of features. The percentiles can vary significantly depending on the features. Table 3 supports the following colloquial statements for potential UAS operations in California:

- Minimum: "Railways and roads are sometimes co-located."
- Mean: "On average along a railway, a UAS comes within 1.93 nm of a road but comes within 26.97 nm of a wind turbine."
- Median: "At any given point along a railway, a UAS comes usually within 0.61 nm of a road."

- Maximum: "The closest point of approach between UAS inspecting a railway and road may exceed 60 nm."

**Table 3.** Closet point of approach percentiles for US-CA across all feature combinations. Means are nautical miles with extremes highlighted.

| Feature #1 | Feature #2 | Mean | 0 | 5 | 25 | 50 | 75 | 95 | 100 |
|---|---|---|---|---|---|---|---|---|---|
| FAA Obstacles | Golf Course | 5.33 | 0.00 | 0.38 | 1.44 | 2.93 | 6.58 | 21.51 | 56.66 |
| FAA Obstacles | Pipeline | 4.05 | 0.00 | 0.10 | 0.57 | 1.39 | 4.06 | 18.66 | 60.00 |
| FAA Obstacles | Railway | 3.22 | 0.00 | 0.06 | 0.56 | 1.69 | 3.98 | 10.12 | 60.00 |
| FAA Obstacles | Road | 2.96 | 0.00 | 0.08 | 0.54 | 1.58 | 3.95 | 9.88 | 55.61 |
| FAA Obstacles | Wind Turbine | 20.61 | 0.00 | 0.04 | 1.37 | 18.30 | 35.04 | 57.93 | 60.00 |
| Electric Power | FAA Obstacles | 2.85 | 0.00 | 0.05 | 0.57 | 1.60 | 3.72 | 9.82 | 33.07 |
| Electric Power | Golf Course | 7.35 | 0.00 | 0.53 | 1.90 | 4.42 | 10.18 | 23.48 | 44.50 |
| Electric Power | Pipeline | 5.73 | 0.00 | 0.14 | 0.82 | 2.10 | 5.92 | 23.18 | 60.00 |
| Electric Power | Railway | 5.44 | 0.00 | 0.08 | 0.90 | 2.72 | 7.13 | 20.52 | 60.00 |
| Electric Power | Road | 3.50 | 0.00 | 0.08 | 0.65 | 1.83 | 4.47 | 13.13 | 29.27 |
| Electric Power | Wind Turbine | 29.18 | 0.00 | 3.49 | 14.19 | 27.07 | 42.63 | 60.00 | 60.00 |
| Golf Course | Pipeline | 3.29 | 0.00 | 0.12 | 0.58 | 1.45 | 3.12 | 12.92 | 60.00 |
| Golf Course | Railway | 3.50 | 0.00 | 0.26 | 1.01 | 2.03 | 4.18 | 12.58 | 60.00 |
| Golf Course | Road | 2.27 | 0.00 | 0.15 | 0.66 | 1.42 | 2.97 | 6.58 | 55.68 |
| Golf Course | Wind Turbine | 26.21 | 0.36 | 5.22 | 12.05 | 22.98 | 37.95 | 60.00 | 60.00 |
| Pipeline | Railway | 4.92 | 0.00 | 0.13 | 0.87 | 2.47 | 6.56 | 18.95 | 34.74 |
| Pipeline | Road | 3.88 | 0.00 | 0.12 | 0.71 | 1.96 | 5.08 | 13.52 | 34.63 |
| Pipeline | Wind Turbine | 30.03 | 0.00 | 3.56 | 15.04 | 28.19 | 43.44 | 60.00 | 60.00 |
| Railway | Road | 1.93 | 0.00 | 0.03 | 0.20 | 0.61 | 1.53 | 10.20 | 40.38 |
| Railway | Wind Turbine | 26.97 | 0.02 | 3.33 | 11.50 | 25.04 | 39.14 | 60.00 | 60.00 |
| Road | Wind Turbine | 34.11 | 0.01 | 4.49 | 18.29 | 33.01 | 53.51 | 60.00 | 60.00 |

Figures 2 and 3 illustrate the distribution of various closet points of approach percentiles across all locations and features. Similar to the California-only results, different features are often co-located or within one nautical mile apart across all locations. The majority of the medians were less than 2.6 nautical miles and the majority of the 25th percentiles were about 1 nautical mile or less. Wind turbines often were the farthest away from other features, while roads were the most often near other features.

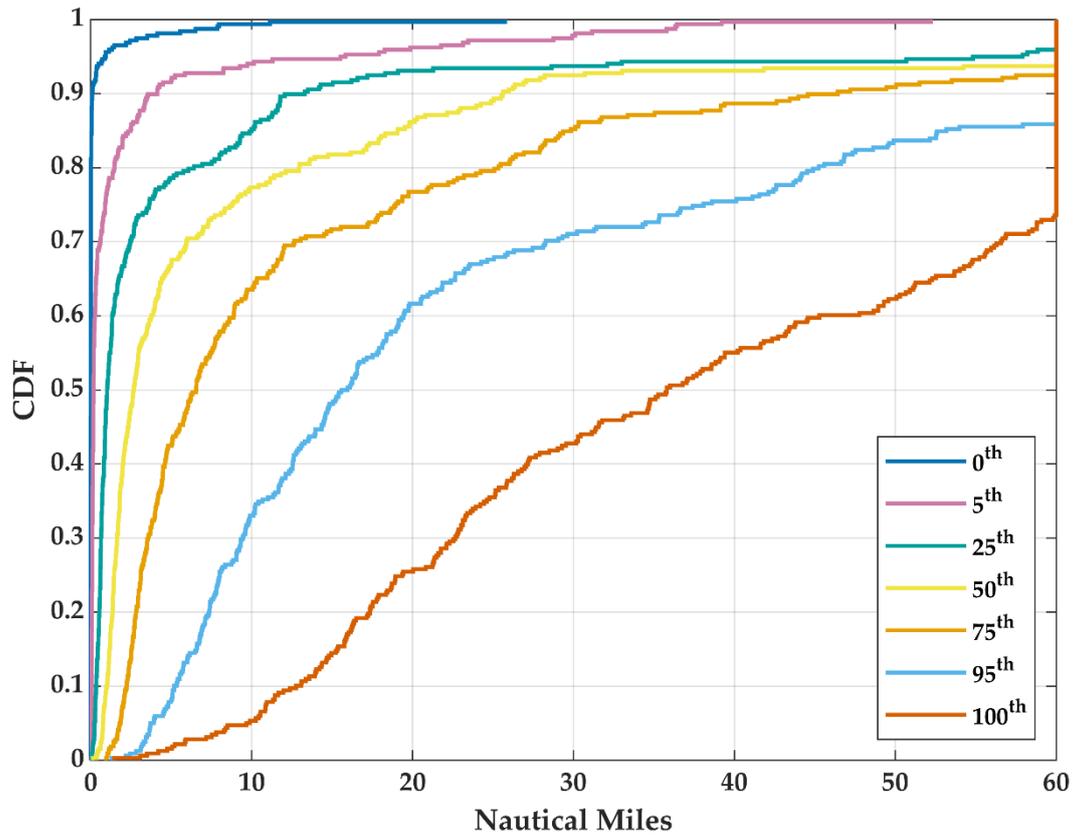

**Figure 2.** Closest point of approach percentile distributions across all locations and features.

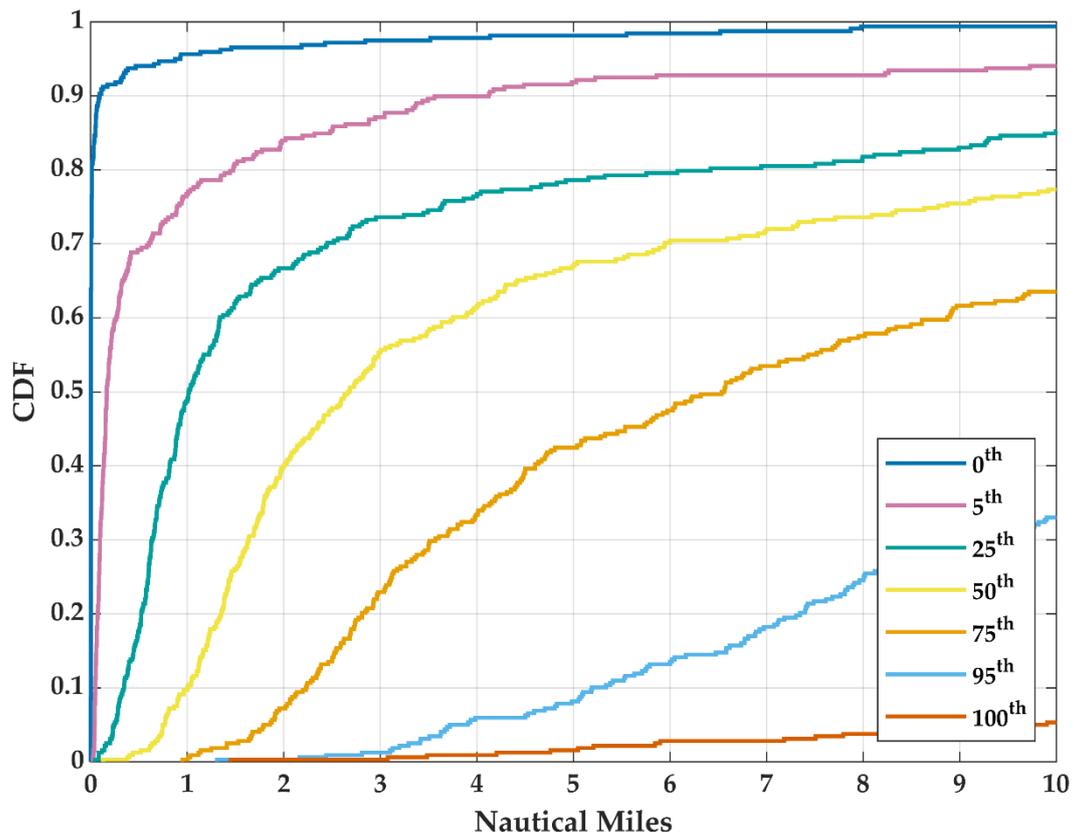

**Figure 3.** Percentile distributions across all locations and features; figure restricted to ten nautical miles.

## 4. Discussion

This section starts with a brief discussion on the sensitivity of the results to using a closer interpolation and then illustrates how geography or urban planning influenced the results. This section concludes with an example of how data availability and coverage can skew the results.

*4.1. Results Sensitvity to Spacing between Points*

To assess the sensitivity of the results to the spacing between interpolated points, we repeated the analysis for one of the largest USA states, California, using a 100 instead of a 500 foot spacing. Shown by Table 4, the percent difference (given two numbers, a and b), for the computation time, mean, and median was calculated as

$$\% \text{ Difference} = 100 \times (a - b)/b. \qquad (1)$$

There was mostly a negligible difference between the mean and median statistics when comparing the use of 100 and 500 feet. However, there was a significant percent increase in the required computation time due to the increased quantity of points to consider. The trade-off for reducing the interpolated spacing between points did not provide a sufficient incentive to repeat the analysis for all locations.

Table 4. Percent differences for US-CA results when changing spacing from 500 to 100 feet.

| Feature #1 | Feature #2 | Total Point Pairs | Compute Time (s) | Mean (nm) | Median (nm) |
|---|---|---|---|---|---|
| FAA Obstacles | Golf Course | 1754% | 2113% | 4.2% | 6.1% |
| FAA Obstacles | Pipeline | 1797% | 2210% | 4.0% | 1.0% |
| FAA Obstacles | Railway | 1455% | 2623% | 1.0% | 4.4% |
| FAA Obstacles | Road | 1800% | 2705% | 3.3% | 7.3% |
| FAA Obstacles | Wind Turbine | 282% | 2442% | −3.8% | −4.9% |
| Electric Power | FAA Obstacles | 1791% | 356% | 0.3% | 0.6% |
| Electric Power | Golf Course | 2303% | 1681% | 0.3% | 0.6% |
| Electric Power | Pipeline | 2359% | 1826% | 0.2% | 0.4% |
| Electric Power | Railway | 1916% | 2001% | 0.4% | 0.8% |
| Electric Power | Road | 2363% | 1909% | 0.3% | 0.6% |
| Electric Power | Wind Turbine | 395% | 280% | 0.1% | 0.1% |
| Golf Course | Pipeline | 2311% | 2179% | −0.3% | −0.2% |
| Golf Course | Railway | 1876% | 2699% | 0.1% | 0.4% |
| Golf Course | Road | 2315% | 2383% | 0.1% | 0.3% |
| Golf Course | Wind Turbine | 385% | 389% | 0.0% | −0.1% |
| Pipeline | Railway | 1922% | 2712% | 0.2% | 0.2% |
| Pipeline | Road | 2371% | 2461% | 0.2% | 0.2% |
| Pipeline | Wind Turbine | 397% | 346% | 0.1% | 0.2% |
| Railway | Road | 1925% | 2057% | 3.9% | 3.4% |
| Railway | Wind Turbine | 307% | 338% | 0.4% | -0.3% |
| Road | Wind Turbine | 397% | 388% | 0.0% | 0.0% |

410 of 15*4.2. Geography*

The long linear infrastructures of railways, major roads, pipelines, and electric power lines are often no more than a few nautical miles apart. These features are not uniformly distributed across the environment, rather geographical features of mountains and lakes influence the location and man-made features. This is exemplified by Figure 4 for geographical and man-made features around the White Mountains National Forest.

Foremost, the long linear infrastructure traverses through the natural mountain pass south of Franconia Notch near Woodstock. It is more efficient to build power lines and roads through valleys instead of steep and varied mountainous terrain. The mountain pass acts as a natural constraint on the location of the manmade features. The electric power lines to the west of the mountains were likely constructed there to minimize costly construction and maintenance due to challenging terrain. A different constraint is Lake Winnipesaukee to the south of the mountains, there are no long linear infrastructure features traversing through it. Instead the railway and roads border the western edge of the lake. For potential UAS operations, this geography will likely increase or decrease the likelihood that two UAS will encounter each other.

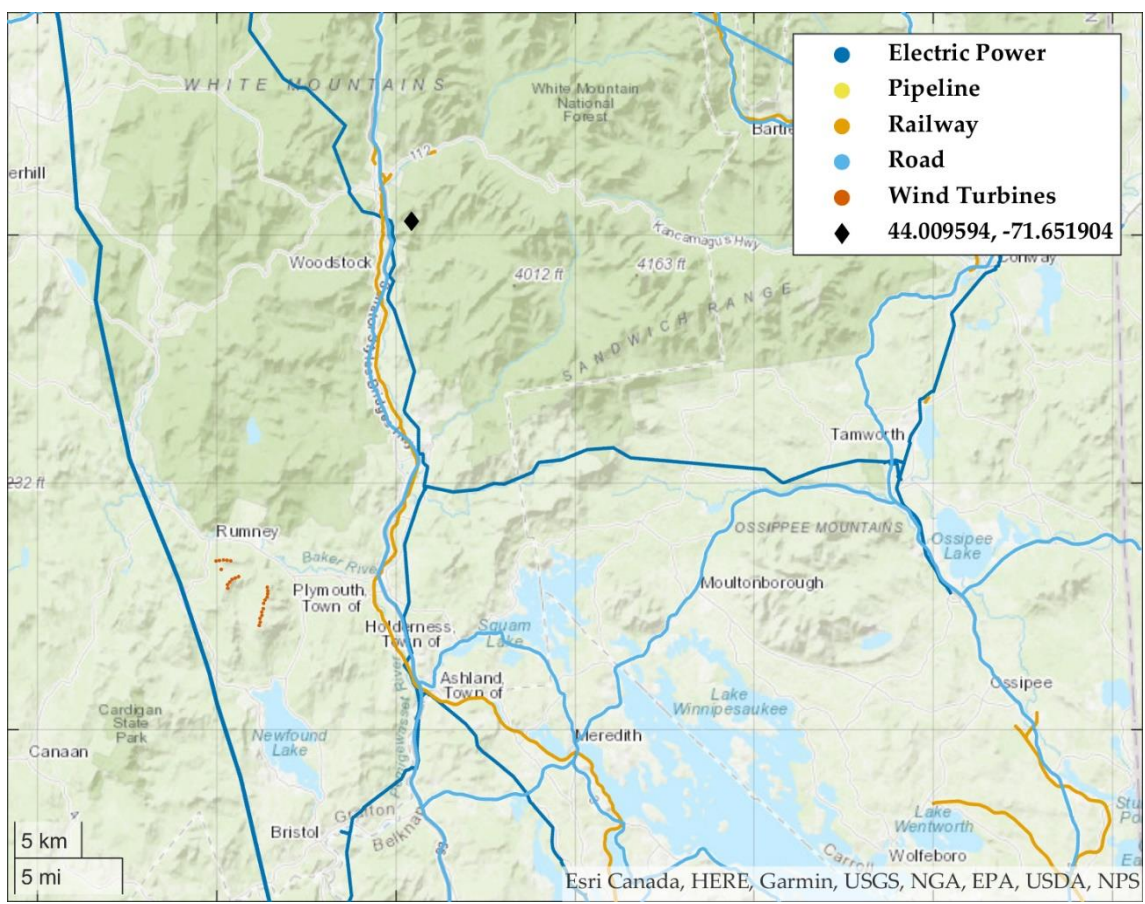

**Figure 4.** Features around the White Mountains in northern New Hampshire.

Nevertheless, as illustrated by Figure 5, geographical features are not solely responsible for influencing the location of man-made features. The environment in Figure 5 is relatively flat and consistent, yet the long linear infrastructures are within 0.5 nautical mile or less of each other. These features align with U.S. Route 2 which was built in 1926, followed by the Great Northern Railway (the eventual Amtrak "Empire Builder") route in 1929. There are engineering design standards to reduce the environmental impact of new transportation routes by locating railway tracks alongside a highway. The nearby transportation system minimizes environmental impact by jointly using land while also increasing the efficiency of multi-vehicle trips.

Additionally, Figure 5 illustrates that specific features, such as wind turbines, are located at specific locations to optimize their operations. The wind turbines are located to maximize energy production while minimizing environmental costs. UAS operations could leverage this when designing risk mitigations. UAS inspections of wind turbines maybe a good candidate to prototype strategic mitigations because our results indicate that they are often far away from other features.

Furthermore, while out of scope for this analysis, these results can inform potential ground collision risk and mitigations. For example, consider that wind turbines are often many miles away from major roadways while railways are often close to major roadways. If a UAS had a critical failure, the likelihood of a UAS crashing into manned vehicles on the road is potentially less if the UAS was inspecting a wind turbine than railway.

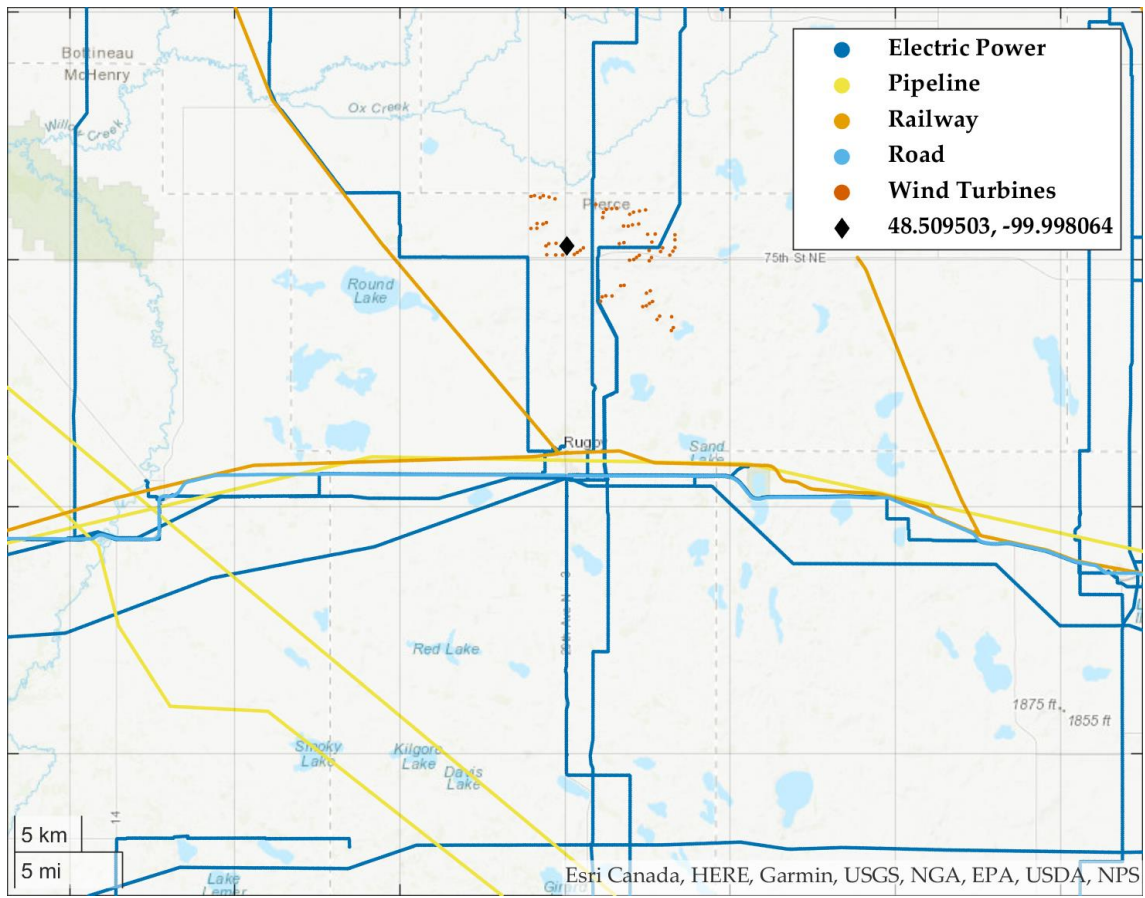

**Figure 5.** Features near Rugby in northern North Dakota.

*4.3. Urban Planning*

Urban planning and the presence of urban clusters or developed land will also influence the interaction between UAS operations. Figure 6 illustrates railways and golf courses near Boston, MA. The railways are more prevalent in the urban center with individual lines closer to each other. As the railways navigate outside the major city, the lines become more dispersed while still serving more developed regions. Conversely, recreational-focused golf courses are located in less developed regions and none exist in Boston's city center. Since railways and golf courses serve different societal needs, it is not surprising that golf courses are railways are often miles apart. Additionally, golf courses are often accessed primarily by roads, rather than other modes of transportation. This reflects that the potential closest point of approach between golf course and roadway inspection UASs are closer than those between golf course and railway operations.

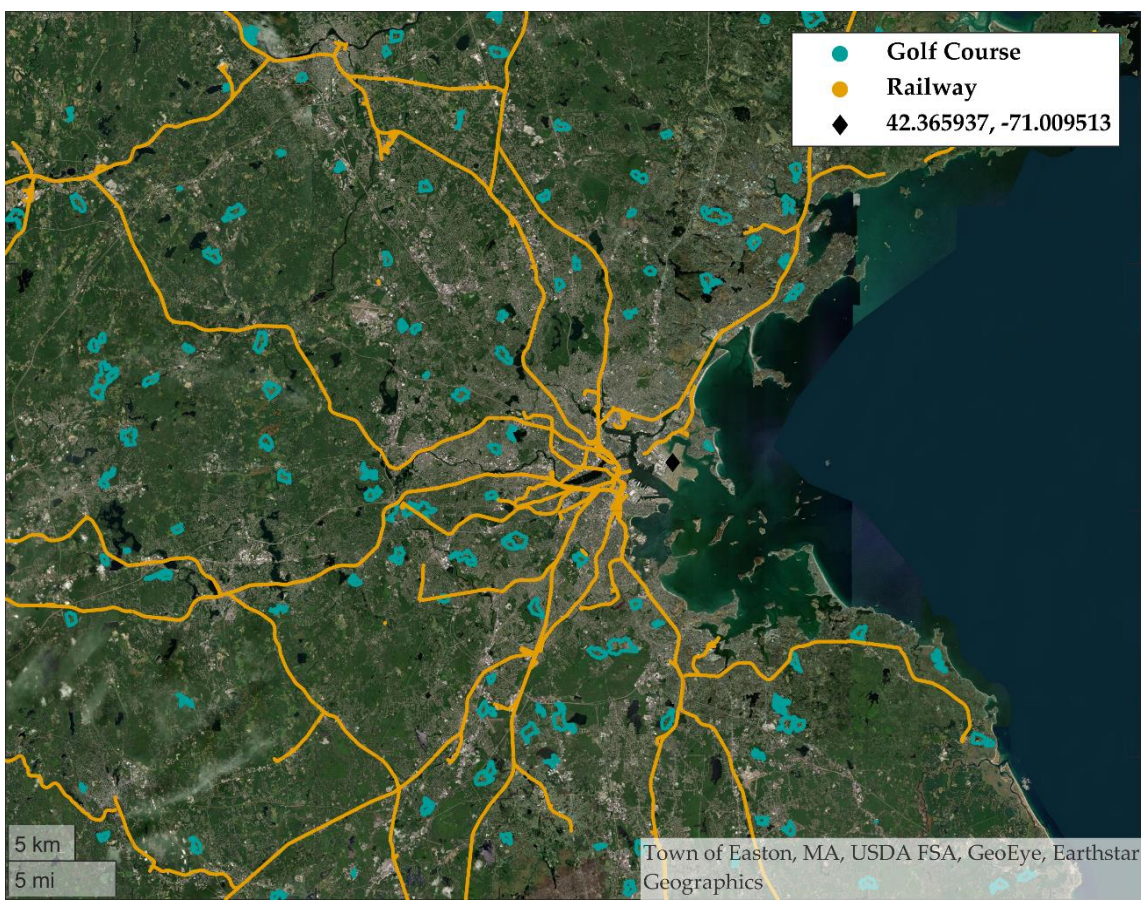

**Figure 6.** Features near Boston in eastern Massachusetts.

*4.4. Data Availability and Coverage*

Lastly, Figure 7 illustrates electric power lines, roads, and FAA obstacles in southern Rhode Island. The electric power line is parallel to the road and the FAA reported obstacles are co-located with the power line. These obstacles are the towers supporting the power lines themselves. As described in Section 2.3, each obstacle is represented as a circle with a radius defined by the horizontal position uncertainty. The larger the circle, the greater the uncertainty. A challenge is that many datasets do not guarantee complete coverage of all features. This issue exists for both federally managed and open sourced datasets. The information available across datasets varies too and correlating datasets can be challenging.

In Figure 7, the DHS HIFLD electric power line datasets do not specify the locations of the towers supporting the lines. The FAA DOF simply specifies the location of a tower and does not designate if a tower is used to support electric power lines. However, from a UAS operations perspective, an inspection of electrical systems could include both the tower and power lines. The consequence is that our results may skew towards closer smaller distances, as we do not delineate if two features are components of a single system.

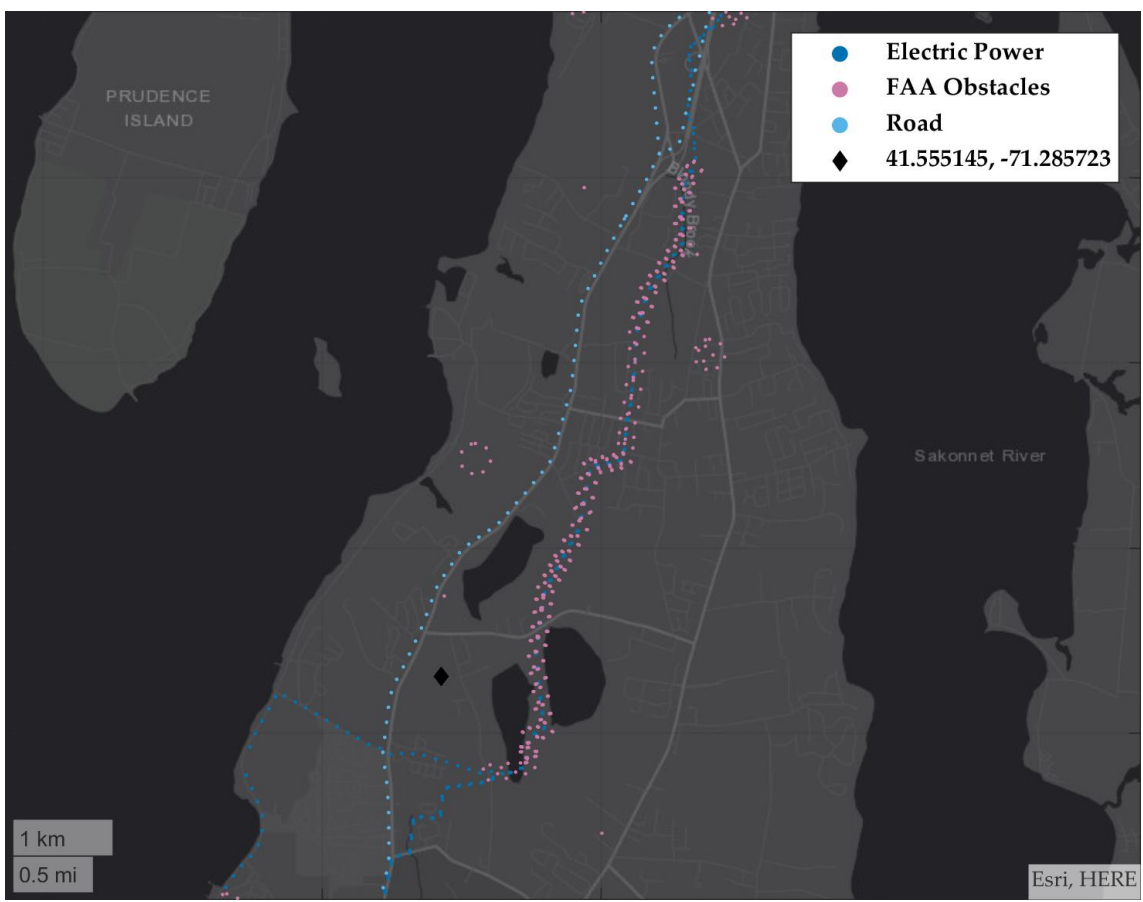

**Figure 7.** Features near Portsmouth in southern Rhode Island.

## 5. Conclusions

We demonstrated an analytical method to characterize potential UAS encounters. By characterizing encounters, we can develop and evaluate systems that mitigate airborne collision risk. These system could be obstacle aware, like ACAS sXu [10], such that when maneuvering to avoid another aircraft, the system will attempt to not inadvertently maneuver into an obstacle. Representing obstacles as part of the simulation would be beneficial for system development, evaluation and deployment. The systems could also be cooperative in that aircraft exchange information via an airborne vehicle to vehicle link; and this link would need to be considered as part of DAA Monte Carlo simulations.

However, ASTM [9,19] nor RTCA have published standards that provide requirements on how to represent obstacles or performance requirements on a V2V system, as part of a DAA system for smaller UAS. These DAA capabilities are currently being drafted and developed by the standards developing organizations and civil aviation authorities. The presented research supports these standard activities by determining which obstacles should be prioritized for inclusion in DAA simulations; assessing if specific encounter scenarios are suitable to assess a given V2V technology; and characterizing encounter scenarios that are likely rare, regardless of UAS flight hours.

For example, wind turbines were consistently many miles away from the other features. Except for when considering FAA obstacles and wind turbines, the $5^{th}$ percentile distance between a wind turbine and another feature was at least three nautical miles. In comparison, the $25^{th}$ percentile distance between electrical power transmission lines and other features was less than one nautical mile. Since wind turbines are often much farther away from other features, it is less likely that any potential encounter between smaller UAS would occur near a wind turbine, whereas, simply due to the geospatial distribution of features, encounters are more likely to occur near electrical power transmission lines. This example suggests that there is a greater need to represent electrical power lines in DAA simulations than wind turbines. Additionally, if an obstacle database was stored onboard a UAS and given a specific concept of operations, storing locations of wind turbines may not be an efficient use of limited storage resources.

Likewise, the results can inform if a given V2V link would support different encounter scenarios. While there are many other factors when characterizing a V2V link, such probability of reception, update rate, available spectrum, and transit power, we can illustrate how the presented results can be leveraged with respect to expected transmit range. Specifically, the ASTM RemoteID standard [9] discusses that Bluetooth 4 would have a range of about 1300 feet or less in a rural environment, Bluetooth 5 Long Range could have a range of 3280 feet or less, and Wi-Fi Aware could have a range of 6561 feet or less. Given that wind turbines are often miles away from other features, regardless of flight hours, the results suggest that a UAS, with a DAA communication range of one nautical mile, inspecting a wind turbine would often not be within range to communicate with other UASs. Accordingly, wind turbine inspections may not be a good use case to drive development of a V2V link for DAA. Conversely, since electric power lines are often closer to other potential UAS operations, a V2V link with a maximum transmit range of one nautical mile maybe enough.

These example analyses are potential future contributions to the aviation safety community and are enabled by our research to quantitatively define a close encounter between UASs. Once UAS operations become more routine or concepts of operations are better established, this research could be extended to consider the frequency of specific encounter scenarios.

**Supplementary Materials:** The following are available with this paper, Table S1: Results for all locations and feature combinations.

**Author Contributions:** All authors have read and agreed to the published version of the manuscript.

**Funding:** Distribution statement A: approved for public release. This material is based upon work supported by the Federal Aviation Administration under Air Force Contract No. FA8702-15-D-0001. Any opinions, findings, conclusions or recommendations expressed in this material are those of the author(s) and do not necessarily reflect the views of the Federal Aviation Administration. Delivered to the U.S. Government with Unlimited Rights, as defined in DFARS Part 252.227-7013 or 7014 (Feb 2014). Notwithstanding any copyright notice, U.S. Government rights in this work are defined by DFARS 252.227-7013 or DFARS 252.227-7014 as detailed above. Use of this work other than as specifically authorized by the U.S. Government may violate any copyrights that exist in this work.

**Acknowledgments:** The authors greatly appreciate the support provided by Sabrina Saunders-Hodge, Richard Lin, Adam Hendrickson, and Bill Oehlschlager from the Federal Aviation Administration. The authors wish to acknowledge the support of their colleagues, Rodney Cole, and Ngaire Underhill. The authors also acknowledge the MIT Lincoln Laboratory Supercomputing Center for providing high-performance computing resources that have contributed to the research results reported within this paper.

**Conflicts of Interest:** The authors declare no conflict of interest. The funders had a role in the decision to publish the results.